\documentclass[pmlr]{jmlr}
\twocolumn
\usepackage{lipsum}
\usepackage{multirow}

\usepackage{booktabs}
  \makeatletter
\def\set@curr@file#1{\def\@curr@file{#1}} 
\makeatother
\usepackage[load-configurations=version-1]{siunitx} 

\usepackage{hyperref}
\usepackage{graphicx}

\usepackage{amsmath, amsfonts}
\usepackage{enumitem}
\usepackage{dsfont}
\usepackage{hyphenat}

\jmlrvolume{140}
\jmlryear{2021}
\jmlrworkshop{AAAI Workshop on Meta-Learning and MetaDL Challenge} 

\title{Advances in MetaDL: AAAI 2021 challenge and workshop}

\author{\Name{Adrian El Baz}\textsuperscript{1} \Email{adrian.el\_baz@ens-paris-saclay.fr} \\
  \Name{Isabelle Guyon}\textsuperscript{1,2} \Email{guyon@chalearn.org} \\
  \Name{Zhengying Liu}\textsuperscript{2} \Email{zhengying.liu@inria.fr}\\
  \Name{Jan N. van Rijn}\textsuperscript{3} \Email{j.n.van.rijn@liacs.leidenuniv.nl} \\
  \Name{Sebastien Treguer}\textsuperscript{1} \Email{streguer@gmail.com} \\
  \Name{Joaquin Vanschoren}\textsuperscript{4} \Email{j.vanschoren@tue.nl} \\
  \addr{
    \textsuperscript{1}{ChaLearn, USA}\\
    \textsuperscript{2}{INRIA and Universit\'e Paris Saclay, France}\\ 
    \textsuperscript{3}{Leiden Institute of Advanced Computer Science, Leiden University, the Netherlands}\\
    \textsuperscript{4}{Eindhoven University of Technology, the Netherlands}\\
  }\\
}

\editors{Isabelle Guyon, Jan N. van Rijn, S\'{e}bastien Treguer, Joaquin Vanschoren}

\begin{document}

\maketitle

\begin{abstract}
To stimulate advances in meta-learning using deep learning techniques (MetaDL), we organized in 2021 a challenge and an associated workshop. This paper presents the design of the challenge and its results, and summarizes presentations made at the workshop. The challenge focused on few-shot learning classification tasks of small images.  Participants' code submissions were run in a uniform manner, under tight computational constraints. This put pressure on solution designs to use existing architecture backbones and/or pre-trained networks. Winning methods featured various classifiers trained on top of the second last layer of popular CNN backbones, fined-tuned on the meta-training data (not necessarily in an episodic manner), then trained on the labeled support and tested on the unlabeled query sets of the meta-test data.
\end{abstract}

\section{Introduction}
The performance of many machine learning algorithms depends highly on the quality and quantity of available data, and (hyper)parameter settings. In particular, deep learning methods, including convolutional neural networks, are known to be `data-hungry' and require properly tuned hyperparameters~\citep{lecun2015deep,sharma2019hyperparameter}. Meta-Learning is a way to address both issues~\citep{hospedales2020meta,vanschoren2018meta}. 
Simple, but effective approaches reported recently include pre-training models on similar datasets.
This way, a good model or good hyperparameters can be pre-determined or previously learned model parameters can be transferred to the new dataset.
As such, higher performance can be achieved with the same amount of data or similar performance with less data (few-shot learning). 

The term Meta-Learning has been around for a long time, and initially mainly addressed the algorithm selection problem~\citep{brazdil2008metalearning}, selecting a good algorithm with adequate \emph{hyperparameters} (e.g., the learning rate or the number of layers of a neural network).
This form of Meta-Learning was often based on meta-features and is applied in various Automated Machine Learning tools, to enhance predictive performance~\citep{Feurer2015}. 
The Open Algorithm Selection Challenge (OASC) is a recent example of a competition that was hosted on this topic~\citep{lindauer2017open,lindauer2019algorithm}.

Recently, a new community emerged that adopted the term Meta-Learning as a way of transfer learning~\citep{vanschoren2018meta}.
This form of meta-learning directly optimizes the \emph{parameters} of a model, e.g., the weights of a neural network. 
Many techniques have been proposed or adapted to enable learning across datasets; these are typically divided into one of the following sub-categories: 
metric-based (e.g., prototypical networks~\citep{snell2017prototypical}, Siamese networks~\citep{koch2015siamese}), model-based (e.g., Meta-Learning with MANNs~\citep{santoro2016meta}, meta networks~\citep{munkhdalai2017meta}) and optimization-based (e.g., LSTM optimizer~\citep{andrychowicz2016learning}, MAML~\citep{MAML}).
These are only a few examples from the large body of scientific papers~\citep{hospedales2020meta,huisman2021survey}. 

As with all thriving fields with many active researchers, it is hard to establish consensus about which techniques have the greatest potential and to make meaningful statements about what is state-of-the-art.  
There are only a few tools that can be used with ease. 
Additionally, there are only a few standardized benchmarks and uniform protocols. 
By organizing the MetaDL competition, we aim to address this issue. 
During the AAAI Conference, we hosted the Workshop on Meta-learning with co-hosted MetaDL competition. 

This paper is organized as follows. 
Section~\ref{sec:workshop} reviews the highlights of the workshop, including presentations of keynote speakers, information regarding competition participants, and accepted papers in the areas of transfer of knowledge in deep learning and algorithm selection. 
Section~\ref{sec:metadl} describes the MetaDL competition.
Section~\ref{sec:baselines} describes the baselines and the winning approaches in the MetaDL challenge.
Section~\ref{sec:conclusions} concludes. 

\section{Workshop highlights}\label{sec:workshop}
We will reflect on both the contributions of the invited keynote speakers and the accepted papers.

\subsection{Keynote Speakers} 
The workshop featured four high-profile keynote speakers. All keynotes can be found on the YouTube channel of our workshop\footnote{See: \url{https://www.youtube.com/channel/UCgiiHLuiWkn-qrnHHsXnm2w}}.
The first keynote speaker, Oriol Vinyals (Google Deepmind), gives an introduction to meta-learning as well as an extensive definition. 
One of the key observations is that good benchmarks in meta-learning are always a moving target.
Currently, the miniImageNet benchmark is popular \citep{vinyals2016matching}, but due to the overlap in classes, one can ask themselves the question of whether determining a certain type of dog (e.g., Husky) is challenging, after many other types of dogs (or generalizations of this class-type, such as mammals) have been seen. 
Notably, he points to the new Meta-Dataset benchmark \citep{triantafillou2020meta}, which challenges practitioners to train further out of distribution. 
After this introduction, he reviews the important components of meta-learning (data, model, loss function, and optimization strategy) and presents a taxonomy of meta-learning solutions (model-based, metric-based, and optimization-based).
One in particular interesting model-based technique is GTP3 (OpenAI), which meta-trains on the whole internet and can translate sentences not seen before. 

The second keynote speaker, Chelsea Finn (Stanford University), makes a strong case for the difference between real-world settings and research settings, wherein real-world scenarios, concept drift plays an important role. 
Before diving into this, she gives an introduction to meta-learning, and real-world examples beyond image classification where meta-learning plays an important role, i.e., robotics, predicting properties/activities molecules, adapt models to new diseases, and land cover classification in different regions.
She argues that there are several solutions to concept drift, i.e., fine-tuning the model on the updated data distribution, or build a structure to solve this problem (e.g., convolutions, which work great when we know the structure and how to build it in).
However, both solutions are not always practical. 
In her talk, she discusses two recent works that address various forms of concept drift.
The first work addresses distribution shift, by using MAML to decompose the weights of a neural network into the equivariant structure and corresponding parameters. This way, the latter can be updated in the inner loop of MAML, retaining the equivariant. 
The second work addressed group shift. Common approaches focus on robustness across various groups, at the expense of average accuracy, whereas when using MAML we can focus on adaptivity, retaining average accuracy. 

The third keynote speaker, Lilian Weng (OpenAI), talks about actual meta-learning challenges in robotics. 
A key aspect of meta-learning is defining a series of tasks to learn and transfer information from to the next task. Rather than defining these tasks beforehand, they could be generated deliberately. One such way is automatic domain randomization (ADR) to create a varied set of tasks, and another is asymmetric self-play, in which one agent tries to make the tasks increasingly harder for another agent. This is expertly shown in an OpenAI project in which a Rubik's cube is solved with a single human-like robot hand. Asymmetric self-play is used here to generate a diverse distribution of training tasks. At test time this policy was able to generalize to many unseen tasks such as setting a table, stacking blocks, and solving simple puzzles. Such memory-augmented models trained on an ever-growing distribution of training environments, either generated by ADR or self-play, can do meta-learning at test time and achieve amazing generalizability in environments or tasks that are not seen during training.

The final keynote speaker, Richard Zemel (University of Toronto), reviews various paradigms of meta-learning in the few shot-learning setting. He then focused on semi-supervised few-shot learning, exploiting `context' (either spatial or temporal). Using a synthesizer of panoramic imagery, they developed a new dataset called `roaming room dataset'. It consists of images of interior scenes, displaying various furniture and objects, some labeled and some not. They developed an online contextualized few-shot learning algorithm (called `agent'), capable of recognizing old classes and learning new ones. The room serves as context, but the agent is not informed on room changes and room labels. To be able to keep track of past contexts, the authors implemented a notion of short, medium, and long-term memory, through replay buffers, weights, and memory models (contextual prototypical memory, build on top of CNNs, RNNs, and prototypical networks). Connections were made by the authors between their memory model and cognitive science. The RNN plays the role of the `working memory' (medium-term), the CNN that of the long-term statistical learning, and the prototypical network that of the semantic memory (short term).

\subsection{Accepted papers}
We categorize the accepted papers into three categories: 
(i) the winning submissions to the competition,
(ii) submissions addressing the design and analysis of algorithms for meta-learning to transfer knowledge in deep learning,
and finally (iii) submissions addressing meta-learning for algorithm selection. 

\subsubsection{Competition Participants}
The top two finishers of the competition were invited to submit their solution to the workshop.
\cite{Chen21} propose MetaDelta, a meta-ensemble approach that leverages the 4 GPUs at their disposal to meta-train 4 meta-learners. 
Each meta-learner consists of a pre-trained encoder on ImageNet. 
These backbones are then fine-tuned with the meta-train dataset data in a non-episodic fashion. Beyond the first place in the competition, in the paper, they present good results on other benchmarks as well. 
\cite{Chobola21} contribute an algorithm that is based on PTMAP \citep{PTMAP}, which leverages optimal transport and power transforms. 
First, a ResNet~\citep{he2016deep} backbone is trained on the meta-train dataset in a non-episodic fashion. 
The classifier head is detached and then the resulting encoder is used to extract support and query features of an episode.
A post-processing step is performed to ensure that these vectors follow a Gaussian-like distribution using a power transformation. 
Finally, the Sinkhorn algorithm is used to estimate class centers. 
In the paper, they present good results on both the miniImageNet~\citep{vinyals2016matching} and CUB benchmarks~\citep{wah2011caltech}. 
Both approaches will be described in more detail further on. 

\subsubsection{Transfer knowledge in deep learning}
Several works focus on the design and analysis of algorithms for meta-learning to transfer knowledge in deep learning. 

\cite{Aimen21} devise a stress test to discover the limitations of few-shot learning methods. By increasing the task complexity, they show that initialization strategies for meta-learning methods such as MAML quickly deteriorate, while approaches that use an optimization strategy such as the Metalearner-LSTM~\citep{Ravi2017} perform significantly better. 
Moreover, a hybrid approach combining an optimization strategy for meta-learning trained in a MAML manner works even better and achieves higher transferability from simple to complex tasks.

\cite{Ayyad21} propose prototypical random walks, an extension of prototypical networks, as a graph-based learning signal derived from unlabeled data.
It aims at maximizing the probability of a random walk that begins at the class prototype, yielding a more discriminative representation, where the embedding of the unlabeled data of a particular class got magnetized to its corresponding class prototype. 
They demonstrate the superiority of prototypical random walks on the Omniglot, miniImageNet, and Tiered-ImageNet datasets. 

\cite{Kuo21} address the issue of catastrophic forgetting, which is a common issue in few-shot learning approaches.
A well-known insight is that this can be addressed by injecting the learner with figures from previous tasks.
The authors build upon this insight, and combine it with MetaSGD, further addressing the catastrophic forgetting and also preventing base learners from overfitting and achieved high performances for all tasks.

\cite{Kvinge21} speculate that metric-based approaches (according to the categorization of Vinyals) will only benefit if the classes that the model is evaluated on (meta-test set), are quite similar to the classes that the model is trained on (meta-train set). 
To this end, they develop a new label set, and demonstrate that their hypothesis is valid.
To address the issue, they develop a new model-type, Fuzzy Simplicial Networks, that addresses this issue.

\cite{Majee21} tackle the problem of few-shot object detection (FSOD) in a real-world, class-imbalanced scenario involving road driving images.
It compares methods from both metric learning and meta-learning on two types of task distributions: only road images and a more real-world open set.
The results show that metric learning outperforms meta-learning in both settings, opening new grounds for few-shot learning studies.

\cite{Mitchell21} address the topic of compositional generalization.
They describe compositional generalization with the example of a human knows how to `walk' and how to `walk quickly', once they learn to `cook', they should be able to understand the concepts `cook quickly' by combining the aforementioned concepts.
The authors conclude that meta-learning can serve as a tool for injecting indictive bias into neural networks, however that significant future work is required. 

\subsubsection{Algorithm Selection}

\cite{Kashgarani21} compare two alternative strategies to leverage a `portfolio' of expert algorithms, which are specialists of certain tasks: selection of the best candidate or parallel runs.
The authors study a particular problem of SAT solving.
In their setting, they use a 32 core machine (with 128 GB RAM), which allows them to run up to 32 algorithms in parallel.
The time limit is 5000  CPU  seconds.
Because memory is shared among algorithms, a larger number of runs fail (per algorithm) when parallelism augments (because of time out or out of memory).
They conclude in favor of selecting the best candidate. 

\cite{Leite21} improve on a previous algorithm they proposed, called Active Testing. The algorithm seeks to recommend an algorithm (workflow) on a new task. The meta-training data consists of a performance matrix of algorithms on previous datasets/tasks. The search for the best algorithm on a new task is initialized by the topmost algorithm returned by the average ranking method. The next algorithm to be tested is selected based on the best-predicted performance gain concerning the current incumbent. The performance gain is estimated by averaging the performance gains obtained from meta-training tasks, weighted by the similarity between the new task and previous meta-training tasks. This method does not use classical dataset characteristics, but rather performance-based similarity between vectors of performance values. The Spearman correlation is used in this process. This has the advantage that it can be dynamically updated as more algorithms get tested on the new task. The authors show that this new strategy leads to improved results.

\cite{Liu21} study under which conditions algorithm recommendation can benefit from meta-learning.
They formally define a meta-distribution, i.e., the distribution of algorithm performances across datasets.
In that sense, algorithms can be seen as independent if their performance is independent of the performances of other algorithms and dependent otherwise.
A formal analysis shows that when algorithms are not independent of one another, a greedy search strategy based on the meta-data of other algorithms can often be optimal.

\cite{Meshki21} propose a novel method of inducing abstract meta-features as latent variables learned by a deep neural network.
While model-based and landmarking induce a learning model for each dataset (meta instances) and extract features from this model, in the proposed approach, a model is induced using the meta-base (all datasets).
By training a deep neural network on the traditional meta-feature space, the authors learn a new latent representation for datasets (abstract meta-features).

\section{The MetaDL competition}\label{sec:metadl}
MetaDL is a few-shot learning competition under tight computational constraints. Indeed, each participant has to train and evaluate their models in 2 hours of wallclock time, using 4 GPU's.\footnote{GPU details are provided in section \ref{sec:CodaLab}} 
This is considered a rather low budget, meaning that participants should design smart solutions to deal with this. 
We introduce the MetaDL problem statement and discuss the different choices we made for the MetaDL competition. 

\subsection{Problem statement}
\cite{MAML} defines the few-shot learning problem as the capacity of a model to adapt to a new task with a few training examples.
An image classification task is associated to a dataset $\mathcal{D} = \{ \mathcal{D}^{tr}, \mathcal{D}^{te} \}$ (where typically $\mathcal{D}^{tr}=\{\mathcal{X}^{tr}, Y^{tr} \}$ and $\mathcal{D}^{te}=\{\mathcal{X}^{te}, Y^{te} \}$ both correspond to labeled examples).
In the few-shot learning literature $\mathcal{D}^{tr}$ and $\mathcal{D}^{te}$ are often referred to as \emph{support set} and \emph{query set}, respectively.
In this context, we define a meta-dataset $\mathcal{M}$ as a collection of such image classification tasks.
We denote by $\mathcal{M}^{tr}$ and $\mathcal{M}^{te}$ the meta-train and meta-test dataset respectively.
Since we deal with a multitude of datasets, we indicate a specific dataset $i$ from the meta-dataset with subscript $\mathcal{D}_i = \{ \mathcal{D}_i^{tr}, \mathcal{D}_i^{te} \}$.
In those cases, one dataset  $\mathcal{D}_i$ is often called an \emph{episode}.
In cases when the context indicates that there is only one dataset, we omit the subscript from the notation.
Still, it is important to note that $\mathcal{D}_i$ and $\mathcal{D}$ represent the same concept.

The essence of the meta-learning problem is to provide a learning procedure that can extract information from tasks in $\mathcal{M}^{tr}$ such that the model can adapt using only a few training examples on a new unseen task from $\mathcal{M}^{te}$. 
The leveraged information from $\mathcal{M}^{tr}$ can take various forms such as weight initialization \citep{MAML}, a learned metric \citep{snell2017prototypical} or even a neural network's update rule \citep{Ravi2017,huisman2021stateless}. 

In the context of our challenge, each \emph{phase} (will be explained further on) consists of the problem defined by a meta-train stage a meta-test stage).
More specifically, a phase is composed of a couple $\{ \mathcal{M}^{tr}, \mathcal{M}^{te}\}$ and participants have to extract relevant information from $\mathcal{M}^{tr}$ to perform well on unseen tasks from $\mathcal{M}^{te}$.
During meta-train, the participants train a meta-learner that seeks to perform well across all datasets from $\mathcal{M}^{tr}$, whereas during meta-test it is supposed to adapt to a specific test-dataset $\mathcal{D}$.

In the context of classical machine learning, we define a learner $L$ as:
\begin{equation}\label{eq:learn}
    L : \mathcal{D}^{tr} \to P
\end{equation}
where $P$ is a predictor $\mathcal{P} : \mathcal{X}^{te} \to \mathcal{Y}^{te}$.
A meta-learner in our context is defined as:
\begin{equation}\label{eq:metalearn}
   \mathcal{A}_{meta}: \mathcal{M}^{tr} \to L
\end{equation}

Therefore the goal is to find $\mathcal{A}_{meta}$ such that the resulting learner $L$ could perform well on a new dataset $\mathcal{D}$ from $\mathcal{M}^{te}$.  More specifically, $L$ is optimized on the associated $\mathcal{D}^{tr}$ and outputs $P$ such that its performance predicting the labels in $\mathcal{D}^{te}$ is high.

In contrast with previous challenges like AutoDL \citep{liu_winning_2020}, we focus on a single modality: images. Moreover, we evaluate meta-algorithms on their performance on tasks composed of unseen image classes, thus emphasizing quick adaptation. 

\subsection{Competition design}

The competition was held online: a participant had to submit a python script containing the definition of their meta-algorithm so that we can run and evaluate their approach from scratch. 
Each submitted script respects a specific API that we defined as challenge organizers. This API is meant to be flexible enough to be used to describe any meta-learning procedure. 
It defines 3 main classes for which their methods need to be overridden to completely define a meta-learning algorithm. Its design relies on the definition of the different algorithms' implementation levels that have been described by \cite{unifiedC}. 
The 3 main classes are the following:

\begin{itemize}
    \item \textbf{Meta-learner}: has a \texttt{meta\_fit()} method that encapsulates the meta-training procedure. Using the previously defined notation, it essentially process the meta-dataset and capture the reusable information across meta-training tasks. It takes $\mathcal{M}^{tr}$ as an argument and outputs a \textbf{learner}.
    \item \textbf{Learner}: has a \texttt{fit()} method that encapsulates the training procedure (e.g., the adaptation phase). It takes $\mathcal{D}^{tr}$ as an argument from a dataset $\mathcal{D}$ along with the associated information 
    from the meta-learning procedure to output a \textbf{predictor}.
    \item \textbf{Predictor}: has a \texttt{predict()} method that predicts the labels of test examples. It takes $\mathcal{D}^{te}$ as an argument for the associated dataset $\mathcal{D}$.
\end{itemize}

\subsubsection{The learning procedure}

Once a participant makes a submission, two stages will be invoked, i.e., the meta-train and meta-test stage. 
In meta-train, the uploaded algorithm is presented with the meta-data and a learner is to be returned (in correspondence with Eq.~\ref{eq:metalearn}).
In the meta-test stage, the returned learner from the previous phase is presented with the train observations from datasets, and a labeling is to be returned (in correspondence with Eq.~\ref{eq:learn}). 
We describe both stages in more detail. 

\begin{figure*}[ht]
    \centering
    \includegraphics[width=9cm]{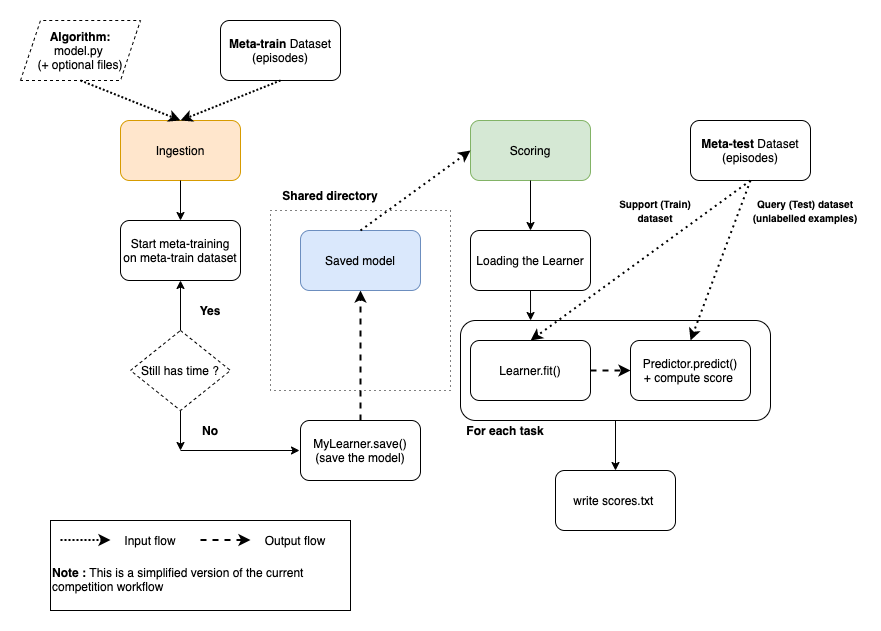}
    \caption{\textbf{MetaDL challenge's evaluation process} for a single phase defined by its associated meta-dataset. The evaluation of a submission is divided into 2 processes: ingestion and scoring. 
    When a participant submits a solution, the ingestion process starts which is essentially the meta-training phase. 
    Once this process terminates, the scoring (i.e., evaluation) process starts. 
    The resulting learner from the ingestion phase is loaded, trained on support data, and evaluated on query examples for each task. 
    This last step is done for 600 episodes. 
    Note that participants need to make sure the 2 processes are completed in 2 hours maximum.}
    \label{fig:metadl-evaluation}
\end{figure*}

\textbf{Meta-training} The ideal few-shot learning algorithm can leverage the knowledge gathered from all these different tasks in $\mathcal{M}^{tr}$ to increase performance on unseen tasks at meta-test time. 
Within a phase, the data are sampled from the associated $\mathcal{M}^{tr}$. 
The way we sample these tasks will be described in section \ref{sec:taskgen}. 

The participants have the choice of selecting a specific configuration of the data fed to their meta-algorithm. There are 2 different types of configurations one can choose from: 
\begin{itemize}
  \item \textbf{Episode mode}: The data arrive in the form of episodes, where each episode represents a given dataset of an equal amount of classes.
  Each of these tasks has the same dataset format that is often used in standard machine learning with the particularity of having just a few examples in the train set.
  \item \textbf{Batch mode}: The data arrive in the form of a batch, completely ignoring the structure of episodes. It could be viewed as a very large standard dataset containing all the classes and their associated examples we could have generated with the episode mode. This technique is usually considered to construct benchmarks and measure the meta-learning effect of approaches that have been trained episodically. 
\end{itemize}

\textbf{Meta-testing} During this phase, the data arrive in the form of episodes. 
To measure the performance across many image classification tasks, 600 episodes are sampled and used for evaluation, as it is usually done in previous few-shot learning benchmarks. 
For each episode, the labels of the test set are hidden. 

For each new task, we use the learner generated from the meta-training part to output a predictor using $\mathcal{D}^{tr}$. 
This process is encapsulated in the learner's \texttt{fit()} function. 
Then, the learner predicts the labels of test examples (i.e., the unlabelled query examples of the associated task) using the resulting predictor. 
The different meta-learning algorithms differ in the way they generate a learner and how they use the information from $\mathcal{M}^{tr}$ coupled with $\mathcal{D}^{tr}_i$ to perform well on $\mathcal{D}^{te}_i$ for a given dataset $\mathcal{D}_i$.

\subsubsection{Evaluation metrics}
While the meta-algorithms can be fed with data arriving in the form of episodes or batch at meta-train time, they are always evaluated with episodes with a fixed setting at meta-test time. This way, we measure the capacity to quickly adapt to a new task from just a few examples.
Performance is measured with a standard classification metric. These episodes follow the 5-way 1-shot classification protocol which essentially means that image classification episodes have 5 classes and $\mathcal{D}^{tr}$ contains only 1 example per class. We use the associated categorical accuracy to evaluate performance, for an image classification task and its associated dataset $\mathcal{D}^{te}_i$:
\begin{equation}
    CatAccuracy = \sum_{y_{j} {\sim} \mathcal{D}_i^{te}} \dfrac{\mathds{1}_{\hat{y}_{j} = y_{j}}}{N}
\end{equation}
where $\hat{y}_{j}$ and $y_{j}$ are respectively the prediction label and the true label associated to the $j^{th}$ test example of the task, $N$ is the total number of examples in $\mathcal{D}_i^{te}$.

To prevent the randomness impact, we run participants' final implementation three times with different seeds and select the one with the lowest performance. Figure \ref{fig:metadl-evaluation} displays the details of the complete evaluation process.

\subsubsection{Competition Phases}\label{sec:Data}
\begin{table*}[ht]
\centering
    \caption{ {\bf MetaDL meta-datasets summary.} For each phase, image classes are split to define a meta-train, meta-validation, and meta-test dataset.
    }
    \resizebox{\textwidth}{!}{
    \begin{tabular}{|c|c|c|c|c|c|c|c|c|}
        \hline\hline
         & \textbf{Domain} &\multicolumn{2}{|c|}{\textbf{Class number}} & \textbf{Sample number} &\multicolumn{3}{|c|}{\textbf{Tensor dimension}} \\
        
         \textbf{Phase} &  & $\mathcal{M}^{tr}$ & $\mathcal{M}^{te}$ & \textbf{per class} & row & col & channel  \\ \hline \hline
         
         Public (Omniglot) & hand-writing  & 964 & 659 & 20 & 28 & 28 & 3 \\ \hline
         Feedback & objects  & 80 & 20 & 600 & 32 & 32 & 3 \\ \hline
         Final & objects  & 85 & 15& 600 & 32 & 32 & 3 \\ \hline

         \hline
    \end{tabular}}
    \label{tab:datasets}
    \vspace{-0.2cm}
\end{table*}
The few-shot image classification literature has significantly developed during the past few years. 
As stated by \cite{doersch2020crosstransformers}, recent research essentially benchmarks adaptability \citep{triantafillou2020meta}.
Indeed, the Meta-Dataset benchmark was introduced to measure the performance of a meta-learner meta-trained on ImageNet data and then evaluated on image classification tasks from other domains. 
Here we focus on the single domain few-shot learning scenario.
For each phase of the challenge, we construct a pair of meta-datasets $\mathcal{M}^{tr}$ and $\mathcal{M}^{te}$ based on a dataset. 
Forming these two sets essentially means splitting the classes from the original dataset. 
For example, the Omniglot dataset is used during the public phase and contains 1623 classes. 
We create a set of 964 classes for meta-training and 659 classes for meta-testing. 
The data used in meta-training and meta-testing is then sampled from examples associated with the corresponding class split. 
This process is done for each of the 3 phases of the MetaDL challenge.  

The choice of datasets for each of the 3 phases is as follows. The description of the data used is summarized in Table \ref{tab:datasets}.

\begin{itemize}
    \item \textbf{Public}: This first phase consists of letting participants have full access to a popular dataset in the few-shot learning literature. 
    For this, we used Omniglot but changed the original class split that was originally proposed to be more challenging as described by \cite{triantafillou2020meta}. 
    This phase enables participants to test the algorithms and analyse their meta-evaluation results locally.
    \item \textbf{Feedback}: The associated dataset is kept hidden from participants. 
    Participants can submit their code and have performance feedback on the meta-test dataset associated with this phase. 
    We selected CIFAR100 \citep{cifar100} but with a class split similar to the one introduced by \cite{TADAM}. This split aims to be more appropriate for a few-shot learning problem, i.e., it avoids mixing concept types between meta-train, meta-validation and meta-test splits. This phase is run on our systems, and each participant can make up to 5 submissions per day.
    \item \textbf{Final}: This phase determines the final ranking of participants, and is run by the organizers behind the scenes.
    The best meta-algorithm obtained during the feedback phase is \emph{ran from scratch} using a new and unseen meta-train and meta-test dataset. 
    The performance on meta-test episodes is the one we use to rank participants at the end of the competition.
    The data we use for this phase are not public so the meta-algorithm needs to blindly adapt to these new images.
\end{itemize}

\subsubsection{Tasks generation}\label{sec:taskgen}
The MetaDL data generation parts are based on Meta-Dataset data pipeline \citep{triantafillou2020meta}.
We generate batches and episodes from the relevant `pools' of classes. This way, we ensure that the meta-learning algorithm is meta-trained with certain classes and evaluated on unseen classes. 

To create a $N$-way $K$-shot episode, we first uniformly sample $N$ classes. 
Then, we sample $K$ examples per class to create $\mathcal{D}_i^{tr}$, and an arbitrary number of examples per class for $\mathcal{D}_i^{te}$. 
Notice that at the meta-train time, the number of examples in $\mathcal{D}^{te}_i$ is not subject to constraints in our setting. 
Participants can choose this number based on their experiments since it has a direct impact on the learning procedure of most meta-learning algorithms. 
Participants can also experiment with different values of $N$, similar to the experiments by \citep{snell2017prototypical}. 

At meta-test time, the number of test examples per episode is fixed and depends on the available examples per class in a dataset. 
Consider the Omniglot dataset used at the public phase, which contains over 1623 classes of 20 examples each.
For each class, we need one example to be used in the train set, leaving 19 examples for the test set. 
To obtain a stable evaluation, we use all these examples. 
As we use the 5-way 1-shot setting, this makes the episodes consist of 5 train episodes and 95 test episodes. 
It is worth noting that the sampling procedure is controlled on the challenge organizer side via setting random seeds.

\subsubsection{CodaLab: platform for running competitions} \label{sec:CodaLab}

As in the AutoDL Challenge \citep{liu_winning_2020}, the MetaDL challenge was run on CodaLab. The CodaLab platform\footnote{\url{https://competitions.codalab.org/}} is a powerful open-source framework for running competitions that involve result or code submission. CodaLab facilitates the organization of computational competitions from the organizers' point of view, as well as provides a rich set of tools (e.g., forum, leaderboard, online submission) to help participants. In MetaDL, participants can submit their codes and receive (almost) real-time feedback on a leaderboard (i.e., during the feedback phase), where the performances of different participants can be compared. More specifically, every submission was run for 2 hours with the same VMs on Azure:  4 Tesla M60 GPU, 224 Go of RAM, and 24 cores.
Note that 2 hours is relatively short for meta-learning approaches, however, due to the small image sizes, we were able to test these techniques in a low-budget setting. 

\section{Baselines, winning approaches, and results}\label{sec:baselines}
 We provided 3 baselines for the competition, 1)~fo-MAML 2)~prototypical networks, and 3)~a transfer learning-based method. The associated code is available publicly online\footnote{See: \url{https://github.com/ebadrian/metadl}}. 
Table \ref{tab:baseline} regroups the baseline results as well as the winning approaches for all the competition phases.

 \begin{table*}[ht]
    \centering
    \caption{{\bf Baseline and results of winning approaches} on public, feedback and final meta-datasets used in MetaDL challenge. Three baseline methods (fo-MAML, prototypical networks, and transfer learning) are applied to all 3 meta-datasets. Performances are 5-way classification accuracy averaged over 600 meta-test episodes along with the $95\%$ confidence interval. Each algorithm is first trained using meta-train data, then evaluated on meta-test 5-way 1-shot episodes. The Meta\_Learners team had an implementation that was not suited to deal with the high number of classes in Omniglot.}
    \resizebox{\textwidth}{!}{
    \begin{tabular}{|c|c|c|c|c|c|c|}
        \hline
        \multicolumn{7}{|c|}{Method : Accuracy (\%) $\pm$ \ Confidence (\%)} \\
        \hline
        & \multicolumn{3}{|c|}{\textbf{Baselines}} & \multicolumn{3}{|c|}{\textbf{Winning approaches}} \\
         \textbf{Phases} & fo-MAML & Proto Networks & Transfer & Meta\_Learners & ctom & MIG\_Edinburgh 
         \\\hline \hline
         
          Public & 26.4 $\pm$ \ 0.68 & 78.3 $\pm$ \ 0.81 & 27.4 $\pm$ \ 0.77& -- & 89.5 $\pm$ \ 0.75& 69.6 $\pm$ \ 0.87\\ \cline{1-7}
          Feedback & 37.8 $\pm$ \ 0.64& 35.3 $\pm$ \ 0.62 & 21.2 $\pm$ \ 3.7& 65.4 $\pm$ \ 1.02& 48.7 $\pm$ \ 0.83& 40.6 $\pm$ \ 0.63\\ \cline{1-7}
          Final & 25.2 $\pm$ \ 0.52& 26.3 $\pm$ \ 0.56 & 23.2 $\pm$ \ 0.36 & 40.2 $\pm$ \ 0.95 & 35.7 $\pm$ \ 0.85& 28.9 $\pm$ \ 6.3\\ \cline{1-7}
         \hline
         
    \end{tabular}}
    \label{tab:baseline}
    \vspace{-0.2cm}
\end{table*}

\subsection{fo-MAML}
The fo-MAML is the first-order approximation of the popular MAML algorithm described by \citep{MAML}.
It essentially means that we do not involve computing second-order gradient in the bi-level optimization loop.
The model is meta-trained episodically in the 5-way 1-shot classification protocol.
Each meta-training episode has 5 classes, the support set contains 1 example per class and the query set contains 19 examples per class.
The image size considered is 28x28.
We use the 4 convolutional layers presented in the original paper and SGD as the optimization algorithm for both meta-learner's weights and the learner's weight.
We use a learning rate of 0.005 for the former and 0.05 for the latter.
We run the algorithm for 1100 epochs with a meta-batch size of 32 (i.e., the number of tasks considered per outer-loop).
 
 \subsection{Prototypical Networks}
\citet{snell2017prototypical} introduced prototypical networks. This algorithm falls in the category of non-parametric approaches.
Consider a neural network $f_{\theta}$ parametrized by its weights $\theta$, that serves as an embedding function of images and outputs a $d$-dimensional vector. The algorithm starts to compute a prototype for each class via the examples in $\mathcal{D}^{tr}_i$ in the embedding space induced by $f_{\theta}$. Then, to classify an example from $\mathcal{D}^{te}$, it computes the softmax distribution of its distances (e.g., Euclidean distance) with each prototype and assigns the label of the closest one.  

As in the original paper, we use episodes that contain 60 classes and 1 example per class in the meta-training phase. The basic architecture is 4 convolutional layers as in the original paper, taking images of size 28x28 as inputs. We run the algorithm for 1500 epochs with an Adam optimizer~\citep{kingma_adam:_2015} with a learning rate of 0.005. The classifier's distance used is the Euclidean distance. This method is considered an effective approach for few-shot learning image classification due to its simplicity and computational efficiency. Based on our experiments, prototypical networks are generally faster to learn than an optimization-based algorithm and thus require less computational time than counterpart methods. 

 \subsection{Transfer baseline}
The transfer baseline is similar to standard fine-tuning methods. It uses a pre-trained network on miniImageNet and we fine-tune the network with the data in the meta-training phase. In contrast with the previous methods, data arrive in the form of batch during the meta-training phase. Each batch contains 50 images from each of the 
available classes. 
During the meta-testing, the data still comes in the form of episodes as for all few-shot learning algorithms. Therefore we drop the classifier on top of the network and only keep the embedding function.
A new classifier is then trained with the data from an episode support set while the embedding function remains frozen.
We train on these data for 10 epochs using an Adam optimizer with a learning rate of 0.001.

The transfer baseline struggles to obtain a good representation of images at meta-test time and it is probably due to the lack of associated training examples (1 per class). Also, the MAML baseline is slow compared to the prototypical one and it is difficult to obtain an efficient learner within the time limit. Prototypical networks learn faster than the aforementioned methods as it is suited to tackle different tasks and learns a simple metric. Finally, we believe that better results could be obtained with careful hyperparameter tuning.

\subsection{Winning Approaches}
We briefly present the MetaDL top-3 approaches in this section. All the associated code is available on the dedicated website\footnote{https://metalearning.chalearn.org/}. The first 2 teams wrote a paper describing their methods. 
\begin{itemize}
    \item 1st place. Team \textbf{Meta\_Learners} \citep{Chen21}: They proposed MetaDelta, a meta-ensemble approach that leverages the 4 GPU at their disposal to meta-train 4 meta-learners. Each meta-learner consists of a pre-trained encoder on miniImageNet (ResNet50 \citep{he2016deep}, ResNet152 \citep{he2016deep}, WRN50 \citep{zagoruyko2016wide} and MobileNet \citep{howard2017mobilenets}). These backbones are then fine-tuned with the meta-train dataset data in a non-episodic fashion (e.g., 100-way classification). To predict query labels of a meta-valid episode, a parameter-free decoder computes softmax of distances between prototypes and query features (generated from the 4 encoders) as in the prototypical network's algorithm. The `meta-module', which consists of a vote of 4 classifiers taking the concatenation of the 4 softmax distributions, is trained on meta-validation data. The best model is kept to evaluate the system performance with meta-test episodes and use a variant of the aforementioned parameter-free decoder that uses query features to compute prototypes \citep{kye2020meta}. 
    The system also has a central controller to ensure time and resource efficiency. 
    
    \item 2nd place. Team \textbf{ctom} \citep{Chobola21}: Their approach is heavily based on the recent work of \cite{PTMAP} that leverages the optimal transport and power transforms. First, a ResNet backbone is trained on the meta-train dataset in a non-episodic fashion. The classifier head is detached and then the resulting encoder is used to extract support and query features of an episode. A post-processing step is performed to ensure that these vectors follow a Gaussian-like distribution using a power transformation. Finally, the Sinkhorn algorithm is used to estimate class centers.
    The predicted label of a query example is determined by the closest class center to its feature representation. 
    \item 3rd place. Team \textbf{MIG\_Edinburgh}:  Their model is a ConvNet with 3 convolutional layers coupled with a QDA classifier on top of the network. It is trained on the meta-train dataset in an episodic fashion. Performance is evaluated with meta-test episodes with the resulting model. 
    
\end{itemize}

\section{Conclusion and Further Work}\label{sec:conclusions}
We presented the design of a new challenge to stimulate the few-shot learning community to embrace deep learning and tackle the hard problems of automating the learning procedure for few-shot image classification tasks. 
We have run baseline methods and provided a benchmark, under a hard time constraint, of popular methods along with MetaDL winning approaches. 
The setting allowed participants to use existing architecture backbones and/or pre-trained networks, which were a starting point to the solution of all top-ranking participants. Such backbones (pre-trained or not) were (re-)trained on the meta-train dataset, in a non-episodic manner for the top two ranking teams. The classification method then varied from team to team but generally used the output of the second last layer of the backbone as input to a classifier trained on the support set and unlabeled query set. The conclusions may have been biased by the specifics of our setting.
The competition series will continue with an extended version of the MetaDL challenge at NeurIPS 2021, featuring a larger variety of datasets, bigger images, more data, and more computational resources, to consolidate these findings.

\acks{This project received support from Microsoft and ChaLearn. We gratefully acknowledge the support of Microsoft Azure with the Azure credits which were required to create GPU instances used for this research. This work is partially supported by ICREA under the ICREA Academia program. We would like to thank our numerous collaborators for their help and support. This research was partially supported by TAILOR, a project funded by EU Horizon 2020 research and innovation program under GA No. 952215, and ANR Chair of Artificial Intelligence HUMANIA ANR-19-CHIA-00222-01.}

\bibliography{main,MetaDL}

\begin{thebibliography}{43}
\providecommand{\natexlab}[1]{#1}
\providecommand{\url}[1]{\texttt{#1}}
\expandafter\ifx\csname urlstyle\endcsname\relax
  \providecommand{\doi}[1]{doi: #1}\else
  \providecommand{\doi}{doi: \begingroup \urlstyle{rm}\Url}\fi

\bibitem[Aimen et~al.(2021)Aimen, Sidheekh, Madan, and Krishnan]{Aimen21}
Aroof Aimen, Sahil Sidheekh, Vineet Madan, and Narayanan~C. Krishnan.
\newblock Stress testing of meta-learning approaches for few-shot learning.
\newblock In \emph{AAAI Workshop on Meta-Learning and MetaDL Challenge}, volume
  140 of \emph{PMLR}, pages 38--44, 2021.

\bibitem[Andrychowicz et~al.(2016)Andrychowicz, Denil, Colmenarejo, Hoffman,
  Pfau, Schaul, Shillingford, and de~Freitas]{andrychowicz2016learning}
Marcin Andrychowicz, Misha Denil, Sergio~G\'{o}mez Colmenarejo, Matthew~W.
  Hoffman, David Pfau, Tom Schaul, Brendan Shillingford, and Nando de~Freitas.
\newblock Learning to learn by gradient descent by gradient descent.
\newblock In \emph{Advances in Neural Information Processing Systems 29},
  NIPS'16, pages 3981--3989. Curran Associates Inc., 2016.

\bibitem[Ayyad et~al.(2021)Ayyad, Li, Muaz, Albarqouni, and Elhoseiny]{Ayyad21}
Ahmed Ayyad, Yuchen Li, Raden Muaz, Shadi Albarqouni, and Mohamed Elhoseiny.
\newblock Semi-supervised few-shot learning with prototypical random walks.
\newblock In \emph{AAAI Workshop on Meta-Learning and MetaDL Challenge}, volume
  140 of \emph{PMLR}, pages 45--57, 2021.

\bibitem[Brazdil et~al.(2008)Brazdil, Carrier, Soares, and
  Vilalta]{brazdil2008metalearning}
Pavel Brazdil, Christophe~Giraud Carrier, Carlos Soares, and Ricardo Vilalta.
\newblock \emph{Metalearning: Applications to data mining}.
\newblock Springer Science \& Business Media, 2008.

\bibitem[Chen et~al.(2021)Chen, Guan, Wei, Wang, and Zhu]{Chen21}
Yudong Chen, Chaoyu Guan, Zhikun Wei, Xin Wang, and Wenwu Zhu.
\newblock Metadelta: A meta-learning system for few-shot image classification.
\newblock In \emph{AAAI Workshop on Meta-Learning and MetaDL Challenge}, volume
  140 of \emph{PMLR}, pages 17--28, 2021.

\bibitem[Chobola et~al.(2021)Chobola, Va\v{s}ata, and Kord\'{i}k]{Chobola21}
Tom\'{a}\v{s} Chobola, Daniel Va\v{s}ata, and Pavel Kord\'{i}k.
\newblock Transfer learning based few-shot classification using optimal
  transport mapping from preprocessed latent space of backbone neural network.
\newblock In \emph{AAAI Workshop on Meta-Learning and MetaDL Challenge}, volume
  140 of \emph{PMLR}, pages 29--37, 2021.

\bibitem[Doersch et~al.(2020)Doersch, Gupta, and
  Zisserman]{doersch2020crosstransformers}
Carl Doersch, Ankush Gupta, and Andrew Zisserman.
\newblock Crosstransformers: spatially-aware few-shot transfer.
\newblock \emph{arXiv preprint arXiv:2007.11498}, 2020.

\bibitem[Feurer et~al.(2015)Feurer, Klein, Eggensperger, Springenberg, Blum,
  and Hutter]{Feurer2015}
M.~Feurer, A.~Klein, K.~Eggensperger, J.~T. Springenberg, M.~Blum, and
  F.~Hutter.
\newblock Efficient and robust automated machine learning.
\newblock In \emph{Advances in Neural Information Processing Systems 28},
  NIPS'15, pages 2962--2970. Curran Associates, Inc., 2015.

\bibitem[Finn et~al.(2017)Finn, Abbeel, and Levine]{MAML}
Chelsea Finn, Pieter Abbeel, and Sergey Levine.
\newblock {Model-agnostic Meta-learning for Fast Adaptation of Deep Networks}.
\newblock In \emph{Proceedings of the 34th International Conference on Machine
  Learning}, ICML'17, pages 1126--1135. JMLR.org, 2017.

\bibitem[He et~al.(2016)He, Zhang, Ren, and Sun]{he2016deep}
Kaiming He, Xiangyu Zhang, Shaoqing Ren, and Jian Sun.
\newblock Deep residual learning for image recognition.
\newblock In \emph{Proceedings of the IEEE conference on computer vision and
  pattern recognition}, pages 770--778, 2016.

\bibitem[Hospedales et~al.(2020)Hospedales, Antoniou, Micaelli, and
  Storkey]{hospedales2020meta}
Timothy Hospedales, Antreas Antoniou, Paul Micaelli, and Amos Storkey.
\newblock Meta-learning in neural networks: A survey.
\newblock \emph{arXiv preprint arXiv:2004.05439}, 2020.

\bibitem[Howard et~al.(2017)Howard, Zhu, Chen, Kalenichenko, Wang, Weyand,
  Andreetto, and Adam]{howard2017mobilenets}
Andrew~G Howard, Menglong Zhu, Bo~Chen, Dmitry Kalenichenko, Weijun Wang,
  Tobias Weyand, Marco Andreetto, and Hartwig Adam.
\newblock Mobilenets: Efficient convolutional neural networks for mobile vision
  applications.
\newblock \emph{arXiv preprint arXiv:1704.04861}, 2017.

\bibitem[Hu et~al.(2020)Hu, Gripon, and Pateux]{PTMAP}
Yuqing Hu, Vincent Gripon, and St{\'e}phane Pateux.
\newblock Leveraging the feature distribution in transfer-based few-shot
  learning.
\newblock \emph{arXiv preprint arXiv:2006.03806}, 2020.

\bibitem[Huisman et~al.(2021{\natexlab{a}})Huisman, Plaat, and van
  Rijn]{huisman2021stateless}
Mike Huisman, Aske Plaat, and Jan~N van Rijn.
\newblock Stateless neural meta-learning using second-order gradients.
\newblock \emph{arXiv preprint arXiv:2104.10527}, 2021{\natexlab{a}}.

\bibitem[Huisman et~al.(2021{\natexlab{b}})Huisman, van Rijn, and
  Plaat]{huisman2021survey}
Mike Huisman, Jan~N van Rijn, and Aske Plaat.
\newblock A survey of deep meta-learning.
\newblock \emph{Artificial Intelligence Review}, 2021{\natexlab{b}}.

\bibitem[Kashgarani and Kotthoff(2021)]{Kashgarani21}
Haniye Kashgarani and Lars Kotthoff.
\newblock Is algorithm selection worth it? comparing selecting single
  algorithms and parallel execution.
\newblock In \emph{AAAI Workshop on Meta-Learning and MetaDL Challenge}, volume
  140 of \emph{PMLR}, pages 58--64, 2021.

\bibitem[Kingma and Ba(2015)]{kingma_adam:_2015}
Diederik~P. Kingma and Jimmy Ba.
\newblock Adam: {A} {Method} for {Stochastic} {Optimization}.
\newblock In \emph{International Conference on Learning Representations},
  ICLR'15, 2015.

\bibitem[Koch et~al.(2015)Koch, Zemel, and Salakhutdinov]{koch2015siamese}
Gregory Koch, Richard Zemel, and Ruslan Salakhutdinov.
\newblock {Siamese Neural Networks for One-shot Image Recognition}.
\newblock In \emph{Proceedings of the 32nd International Conference on Machine
  Learning}, volume~37 of \emph{ICML'15}. JMLR.org, 2015.

\bibitem[Krizhevsky(2009)]{cifar100}
Alex Krizhevsky.
\newblock Learning multiple layers of features from tiny images.
\newblock Technical report, University of Toronto, 2009.

\bibitem[Kuo et~al.(2021)Kuo, Harandi, Fourrier, Walder, Ferraro, and
  Suominen]{Kuo21}
Nicholas I-Hsien Kuo, Mehrtash Harandi, Nicolas Fourrier, Christian Walder,
  Gabriela Ferraro, and Hanna Suominen.
\newblock Learning to continually learn rapidly from few and noisy data.
\newblock In \emph{AAAI Workshop on Meta-Learning and MetaDL Challenge}, volume
  140 of \emph{PMLR}, pages 65--76, 2021.

\bibitem[Kvinge et~al.(2021)Kvinge, New, Courts, Lee, Phillips, Corley, Tuor,
  Avila, and Hodas]{Kvinge21}
Henry Kvinge, Zachary New, Nico Courts, Jung~H. Lee, Lauren~A. Phillips,
  Courtney~D. Corley, Aaron Tuor, Andrew Avila, and Nathan~O. Hodas.
\newblock Fuzzy simplicial networks: A topology-inspired model to improve task
  generalization in few-shot learning.
\newblock In \emph{AAAI Workshop on Meta-Learning and MetaDL Challenge}, volume
  140 of \emph{PMLR}, pages 77--89, 2021.

\bibitem[Kye et~al.(2020)Kye, Lee, Kim, and Hwang]{kye2020meta}
Seong~Min Kye, Hae~Beom Lee, Hoirin Kim, and Sung~Ju Hwang.
\newblock Meta-learned confidence for few-shot learning.
\newblock \emph{arXiv preprint arXiv:2002.12017}, 2020.

\bibitem[LeCun et~al.(2015)LeCun, Bengio, and Hinton]{lecun2015deep}
Yann LeCun, Yoshua Bengio, and Geoffrey Hinton.
\newblock Deep learning.
\newblock \emph{nature}, 521\penalty0 (7553):\penalty0 436--444, 2015.

\bibitem[Leite and Brazdil(2021)]{Leite21}
Rui Leite and Pavel Brazdil.
\newblock Exploiting performance-based similarity between datasets in
  metalearning.
\newblock In \emph{AAAI Workshop on Meta-Learning and MetaDL Challenge}, volume
  140 of \emph{PMLR}, pages 90--99, 2021.

\bibitem[Lindauer et~al.(2017)Lindauer, van Rijn, and
  Kotthoff]{lindauer2017open}
Marius Lindauer, Jan~N van Rijn, and Lars Kotthoff.
\newblock Open algorithm selection challenge 2017: Setup and scenarios.
\newblock In \emph{Open Algorithm Selection Challenge 2017}, volume~79 of
  \emph{PMLR}, pages 1--7, 2017.

\bibitem[Lindauer et~al.(2019)Lindauer, van Rijn, and
  Kotthoff]{lindauer2019algorithm}
Marius Lindauer, Jan~N van Rijn, and Lars Kotthoff.
\newblock The algorithm selection competitions 2015 and 2017.
\newblock \emph{Artificial Intelligence}, 272:\penalty0 86--100, 2019.

\bibitem[Liu and Guyon(2021)]{Liu21}
Zhengying Liu and Isabelle Guyon.
\newblock Asymptotic analysis of meta-learning as a recommendation problem.
\newblock In \emph{AAAI Workshop on Meta-Learning and MetaDL Challenge}, volume
  140 of \emph{PMLR}, pages 100--114, 2021.

\bibitem[Liu et~al.(2019)Liu, Xu, Madadi, Junior, Escalera, Rajaa, and
  Guyon]{unifiedC}
Zhengying Liu, Zhen Xu, Meysam Madadi, Julio~Jacques Junior, Sergio Escalera,
  Shangeth Rajaa, and Isabelle Guyon.
\newblock Overview and unifying conceptualization of {Automated} {Machine}
  {Learning}.
\newblock \emph{Automating Data Science workshop @ ECML PKDD 2019}, page~8,
  September 2019.

\bibitem[Liu et~al.(2021)Liu, Pavao, Xu, Escalera, Ferreira, Guyon, Hong,
  Hutter, Ji, Junior, Li, Lindauer, Luo, Madadi, Nierhoff, Niu, Pan, Stoll,
  Treguer, Wang, Wang, Wu, Xiong, Zela, and Zhang]{liu_winning_2020}
Zhengying Liu, Adrien Pavao, Zhen Xu, Sergio Escalera, Fabio Ferreira, Isabelle
  Guyon, Sirui Hong, Frank Hutter, Rongrong Ji, Julio C. S.~Jacques Junior,
  Ge~Li, Marius Lindauer, Zhipeng Luo, Meysam Madadi, Thomas Nierhoff, Kangning
  Niu, Chunguang Pan, Danny Stoll, Sebastien Treguer, Jin Wang, Peng Wang,
  Chenglin Wu, Youcheng Xiong, Arb\"er Zela, and Yang Zhang.
\newblock Winning solutions and post-challenge analyses of the chalearn autodl
  challenge 2019.
\newblock \emph{IEEE Transactions on Pattern Analysis and Machine
  Intelligence}, 43\penalty0 (9):\penalty0 3108--3125, 2021.

\bibitem[Majee et~al.(2021)Majee, Agrawal, and Subramanian]{Majee21}
Anay Majee, Kshitij Agrawal, and Anbumani Subramanian.
\newblock Few-shot learning for road object detection.
\newblock In \emph{AAAI Workshop on Meta-Learning and MetaDL Challenge}, volume
  140 of \emph{PMLR}, pages 115--126, 2021.

\bibitem[Meskhi et~al.(2021)Meskhi, Rivolli, Mantovani, and Vilalta]{Meshki21}
Mikhail~M. Meskhi, Adriano Rivolli, Rafael~G. Mantovani, and Ricardo Vilalta.
\newblock Learning abstract task representations.
\newblock In \emph{AAAI Workshop on Meta-Learning and MetaDL Challenge}, volume
  140 of \emph{PMLR}, pages 127--137, 2021.

\bibitem[Mitchell et~al.(2021)Mitchell, Finn, and Manning]{Mitchell21}
Eric Mitchell, Chelsea Finn, and Chris Manning.
\newblock Challenges of acquiring compositional inductive biases via
  meta-learning.
\newblock In \emph{AAAI Workshop on Meta-Learning and MetaDL Challenge}, volume
  140 of \emph{PMLR}, pages 138--148, 2021.

\bibitem[Munkhdalai and Yu(2017)]{munkhdalai2017meta}
Tsendsuren Munkhdalai and Hong Yu.
\newblock Meta networks.
\newblock In \emph{Proceedings of the 34th International Conference on Machine
  Learning}, ICML'17, pages 2554--2563. JLMR.org, 2017.

\bibitem[Oreshkin et~al.(2018)Oreshkin, Rodriguez, and Lacoste]{TADAM}
Boris~N. Oreshkin, Pau Rodriguez, and Alexandre Lacoste.
\newblock Tadam: Task dependent adaptive metric for improved few-shot learning.
\newblock In \emph{Advances in Neural Information Processing Systems 32},
  NIPS'18, pages 719--729, Red Hook, NY, USA, 2018. Curran Associates Inc.

\bibitem[Ravi and Larochelle(2017)]{Ravi2017}
Sachin Ravi and Hugo Larochelle.
\newblock {Optimization as a Model for Few-Shot Learning}.
\newblock In \emph{International Conference on Learning Representations},
  ICLR'17, 2017.

\bibitem[Santoro et~al.(2016)Santoro, Bartunov, Botvinick, Wierstra, and
  Lillicrap]{santoro2016meta}
Adam Santoro, Sergey Bartunov, Matthew Botvinick, Daan Wierstra, and Timothy
  Lillicrap.
\newblock {Meta-learning with Memory-augmented Neural Networks}.
\newblock In \emph{Proceedings of the 33rd International Conference on
  International Conference on Machine Learning}, ICML'16, pages 1842--1850.
  JMLR.org, 2016.

\bibitem[Sharma et~al.(2019)Sharma, van Rijn, Hutter, and
  M{\"u}ller]{sharma2019hyperparameter}
Abhinav Sharma, Jan~N van Rijn, Frank Hutter, and Andreas M{\"u}ller.
\newblock Hyperparameter importance for image classification by residual neural
  networks.
\newblock In \emph{International Conference on Discovery Science}, pages
  112--126. Springer, 2019.

\bibitem[Snell et~al.(2017)Snell, Swersky, and Zemel]{snell2017prototypical}
Jake Snell, Kevin Swersky, and Richard Zemel.
\newblock {Prototypical Networks for Few-shot Learning}.
\newblock In \emph{Advances in Neural Information Processing Systems 30},
  NIPS'17, pages 4077--4087. Curran Associates Inc., 2017.

\bibitem[Triantafillou et~al.(2020)Triantafillou, Zhu, Dumoulin, Lamblin, Evci,
  Xu, Goroshin, Gelada, Swersky, Manzagol, and
  Larochelle]{triantafillou2020meta}
Eleni Triantafillou, Tyler Zhu, Vincent Dumoulin, Pascal Lamblin, Utku Evci,
  Kelvin Xu, Ross Goroshin, Carles Gelada, Kevin Swersky, Pierre-Antoine
  Manzagol, and Hugo Larochelle.
\newblock {Meta-Dataset: A Dataset of Datasets for Learning to Learn from Few
  Examples}.
\newblock In \emph{International Conference on Learning Representations},
  ICLR'20, 2020.

\bibitem[Vanschoren(2018)]{vanschoren2018meta}
Joaquin Vanschoren.
\newblock Meta-learning: A survey.
\newblock \emph{arXiv preprint arXiv:1810.03548}, 2018.

\bibitem[Vinyals et~al.(2016)Vinyals, Blundell, Lillicrap, Kavukcuoglu, and
  Wierstra]{vinyals2016matching}
Oriol Vinyals, Charles Blundell, Timothy Lillicrap, Koray Kavukcuoglu, and Daan
  Wierstra.
\newblock {Matching Networks for One Shot Learning}.
\newblock In \emph{Advances in Neural Information Processing Systems 29},
  NIPS'16, pages 3637--3645, 2016.

\bibitem[Wah et~al.(2011)Wah, Branson, Welinder, Perona, and
  Belongie]{wah2011caltech}
C.~Wah, S.~Branson, P.~Welinder, P.~Perona, and S.~Belongie.
\newblock {The Caltech-UCSD Birds-200-2011 Dataset}.
\newblock Technical Report CNS-TR-2011-001, California Institute of Technology,
  2011.

\bibitem[Zagoruyko and Komodakis(2016)]{zagoruyko2016wide}
Sergey Zagoruyko and Nikos Komodakis.
\newblock Wide residual networks.
\newblock \emph{arXiv preprint arXiv:1605.07146}, 2016.

\end{thebibliography}

\end{document}